%% file: main.tex
\documentclass[conference]{IEEEtran}

\pdfoutput=1

\input preamble/preamble
\input preamble/macros

\let\oldbibliography\thebibliography
\renewcommand{\thebibliography}[1]{\oldbibliography{#1}
\setlength{\itemsep}{0pt}} 

\begin{document}

\title{\toolname: Privacy Preserving Peer-to-Peer Learning for Infrastructureless Setups}

\author{\IEEEauthorblockN{1\textsuperscript{st} Ioannis Arapakis}
\IEEEauthorblockA{\textit{Telefonica Research} \\
ioannis.arapakis@telefonica.com}
\and
\IEEEauthorblockN{2\textsuperscript{nd} Panagiotis Papadopoulos}
\IEEEauthorblockA{\textit{Telefonica Research} \\
panpap@ics.forth.gr}
\and
\IEEEauthorblockN{3\textsuperscript{rd} Kleomenis Katevas}
\IEEEauthorblockA{\textit{Brave Software} \\
kkatevas@brave.com}
\and
\IEEEauthorblockN{4\textsuperscript{th} Diego Perino}
\IEEEauthorblockA{\textit{Telefonica Research} \\
diego.perino@gmail.com}
}


\maketitle
\input{sections/0_abstract}

\input{sections/1_introduction}
\input{sections/2_related}
\input{sections/3_threat}
\input{sections/4_design}
\input{sections/5_simulations}
\input{sections/6_implementation}
\input{sections/7_conclusion}

\bibliographystyle{abbrv_compact}
\bibliography{main}
\end{document}

%% file: preamble/preamble.tex
\usepackage{booktabs}  
\usepackage{pbox}
\usepackage{epsfig,endnotes}
\usepackage{epstopdf}
\usepackage{xcolor}
\usepackage{colortbl}
\usepackage{graphicx}
\usepackage{xspace}
\usepackage{enumitem}
\usepackage{url}
\usepackage{amsfonts}
\usepackage{amssymb}
\usepackage{amsmath}
\usepackage{breakurl}
\usepackage[utf8]{inputenc}
\usepackage[tableposition=top,font={scriptsize},figurename=Fig., skip=7pt]{caption}  
\usepackage{pgfplots}
\usepackage{tikz}
\usepgfplotslibrary{groupplots}
\usepackage{bbm}
\usepackage{float}
\usepackage{subcaption}
\usepackage{listings}
\usepackage{multirow}
\usepackage{arydshln}



%% file: preamble/macros.tex
\newcommand{\point}[1]{\vspace{0.01in}\par\vspace{0.01in}\noindent{\textbf{#1:} }}

\newfont{\ttlfnt}{phvb8t at 18pt}
\renewcommand\ttlfnt{\fontfamily{phv}\fontseries{b}\fontsize{18}{22}\selectfont}

\newcommand{\toolname}{P{\selectfont\textsuperscript{4}}L\xspace}
\newcommand{\eg}{{e.g.,}\xspace}

\newcommand{\ie}{{i.e.,}\xspace}
\newcommand{\pp}{peer-to-peer\xspace}

\definecolor{Gray}{gray}{0.9}



%% file: sections/0_abstract.tex
\begin{abstract}
Distributed (or Federated) learning enables users to train machine learning models on their very own devices, while they share only the gradients of their models usually in a differentially private way (utility loss). Although such a strategy provides better privacy guarantees than the traditional centralized approach, it requires users to blindly trust a centralized infrastructure that may also become a bottleneck with the increasing number of users.
In this paper, we design and implement \toolname: a privacy preserving \pp learning system for users to participate in an asynchronous, collaborative learning scheme without requiring any sort of infrastructure or relying on differential privacy. Our design uses strong cryptographic primitives to preserve both the confidentiality and utility of the shared gradients, a set of \pp mechanisms for fault tolerance and user churn, proximity and cross device communications. Extensive simulations under different network settings and ML scenarios for three real-life datasets show that \toolname provides competitive performance to baselines, while it is resilient to different poisoning attacks. We implement \toolname and experimental results show that the performance overhead and power consumption is minimal (less than 3mAh of discharge).
\end{abstract}

%% file: sections/1_introduction.tex
\section{Introduction}
\label{sec:introduction}

Traditional Machine Learning (ML) systems are based on a centralized design where all data is transferred, stored and processed in the cloud, and the user endpoints could later fetch the results via API calls. While this approach provides good performance from a ML perspective, it is extremely costly from a system point of view (\ie storage, processing, bandwidth) and the cloud provider becomes the single point failure. Further, it  raises significant privacy concerns, especially considering the sheer volume of the multidimensional and privacy-sensitive user data the modern mobile and IoT devices collect nowadays. 

To mitigate these concerns, the distributed machine learning paradigm of Federated Learning (FL)~\cite{mcmahan2017communication} came to enable mobile devices to produce an ML model without sharing any user data. Instead, in FL, users selectively share small subsets of their models’ key parameters during training.  However, FL requires the coordination of a central server that orchestrates the training, the clients selection, the updates collection, and weights aggregation. This server constitutes a centralized authority that (1) users are forced to blindly trust to protect them from various information leakage attacks~\cite{10.1145/3133956.3134012,melis2018inference}, where malicious participating clients can de-anonymize sensitive training data through observation and isolation of a victim’s model updates, while at the same time (2) such centralized service is a single point of failure, which can become a bottleneck hindering the overall system scalability and performance~\cite{NIPS2017_f7552665}. 

To address these emerging issues, existing approaches either use (1) differential privacy~\cite{bellet2018personalized}, thus decreasing the utility of the shared gradients, or (2) specialized hardware (\eg Trusted Execution Environments~\cite{PPML}). Further, previous \pp learning systems, that do not require a centralized server, heavily rely on blockchain ledgers~\cite{shayan2019biscotti} and interplanetary file system (IPFS)~\cite{pappas2021ipls}, or again use differential privacy~\cite{10.1007/978-3-030-23502-4_10}.

In this paper, we take the next step and design \toolname: a \pp learning system that does not rely either on differential privacy or on external systems (\eg blockchain or IPFS). Specifically, \toolname is a privacy preserving, fully distributed and infrastructure-less \pp learning system that enables users to participate in an asynchronous decentralized collaborative learning scheme. Our goal is for the users to benefit from the wealth of user data that are distributed across the participating devices, while maintaining accurate personalized models. Specifically in \toolname, users train their models locally by using their (privacy-sensitive) data and they create asynchronous, small, collaborative learning synergies with nearby users, via which they share their model's gradients.
\begin{table*}[t]
\centering
    \tiny
    \begin{tabular}{lllllll}
    \toprule
       \bf Approach & \bf Trust Assumption & \bf Architecture & \bf Protection against Leakage & \bf Data Utility & \bf Scalability & \bf Fault Tolerance \\
    \midrule
        ~\cite{7447103}  & Trusted server & Centralized & Distributed SSGD &  Intact & Scalable & No\\ \hline
        ~\cite{10.5555/2997046.2997105} & Untrusted server & Centralized &  HE \& global DP & Degraded & Limited & No \\ \hline
        ~\cite{pmlr-v22-rajkumar12} & Untrusted server & Centralized & Gaussian perturbation & Degraded & Scalable & No \\   \hline
        ~\cite{8456019} & Untrusted server & Centralized & local DP & Very Degraded & Scalable & No \\\hline
        ~\cite{9282830} & Trusted server+peers & Centralized & None & Intact & Scalable & No \\ \hline
        ~\cite{10.1145/3133956.3133982} & Honest-but-curious server & Centralized & MPC (Shamir’s Secrets) & Intact & Limited (many rounds/user) & No \\  \hline
        ~\cite{8241854} & Untrusted server & Centralized & HE & Intact & Scalable & No \\ \hline
        ~\cite{10.1007/978-3-030-41579-2_40}  & Non-colluding servers & Centralized & MPC \& MAC & Intact & Scalable & No \\ \hline
        ~\cite{Wang_2020} & Noise level depends on server trustworthiness & Centralized & local ADMM & Degraded & Scalable & No \\  \hline
        ~\cite{hu2021gfl} & m trusted nodes & Decentralized (IPFS) & Limited (AES+Blockchain) & Intact & Scalable & Yes \\ \hline
        ~\cite{pappas2021ipls} & Trusted peers & Decentralized (IPFS)  & None & Intact & Scalable & Yes \\ \hline
        ~\cite{shayan2019biscotti} & topology for flooding, $>70\%$ honest stake & Decentralized(blockchain)&  DP + MPC (Shamir secrets) & Degraded & Scalable (bootstrap  needed) & Yes \\ \hline
        ~\cite{9252063} & Byzantine peers & Decentralized (p2p) & None & Intact & Scalable & Yes\\ \hline
        ~\cite{10.1007/978-3-030-23502-4_10} & Untrusted peers & Decentralized (LF Topology) & EASGD + local DP & Severely Degraded & Scalable & Pool fails, if  leader fails  \\  \bottomrule
       \rowcolor{Gray}\bf  \toolname & \bf Non-colluding peers & \bf Decentralized (p2p) and infra-less & \bf HE & Intact & \bf Scalable (Sustains  user churn)&\bf Yes\\
    \bottomrule
    \end{tabular}
    \caption{Summary of related works}\vspace{-0.7cm}
    \label{tab:related}
\end{table*}

In \toolname, we utilize partial Homomorphic Encryption (HE) during gradient averaging to preserve both the confidentiality and utility of the shared gradients, while at the same time we eliminate information leakage attacks by malicious participants. In addition, \toolname leverages the proximity and cross-device communication capabilities~\cite{nearby} of mobile devices, thus operating on completely infrastructureless setups: no need for any centralized federation, Public Key Infrastructure (PKI) or Internet connection. 

The contributions of our work are the following:
\begin{enumerate}[leftmargin=14pt,noitemsep,topsep=0pt]
    \item We design and implement \toolname: a Privacy-Preserving fully distributed \pp learning system for infrastructureless setups. In \toolname, users can collaboratively tune their local personalized models without needing to trust any fellow peer or centralized authority, and does not require any third-party infrastructure. 
    
    \item We perform analytical simulations to assess the performance of \toolname under different network settings and ML scenarios, for three real-life datasets. Our findings demonstrate that \toolname can deliver final global models with competitive performance, or even exceed the performance of golden-standard baselines. Moreover, \toolname shows a \textit{natural} resilience to various types of sophisticated poisoning attacks without compromising significantly the performance.
   
    \item We evaluate the system performance of a \toolname prototype and we show the feasibility of our design on Android OS using three popular smartphones under ML models of various weights. Our findings show that \toolname has minimal overheads on CPU and battery (less than 3 mAh).

\end{enumerate}

%% file: sections/2_related.tex
\section{Related Work}
\label{sec:related}

While the traditional ML paradigm requires user data to be transmitted to a centralized entity that is responsible for training models on the large volumes of received data, yet when it comes to  privacy-sensitive data, they should not ever leave the user's device. On the other hand, training models in isolation (based only on locally available data) may result in poorly performing models, if users have insufficient data, or models that don't generalize well.

To mitigate this exact limitation, Federated Learning (FL)~\cite{mcmahan2017communication} came to enable mobile devices to produce an ML model without sharing any user data. 
Apart from the important privacy benefits provided: (1) no user data leave the users' devices, (2) the use of differential privacy protects from inference attacks (but introduces a privacy to utility trade-off), yet the coordinating server constitutes a centralized authority that users are forced to blindly trust. This server also constitutes a single point of failure, which can become a bottleneck for the scalability and overall system performance with the increasing number of users~\cite{NIPS2017_f7552665}.

As summarized in Table~\ref{tab:related}, there are various approaches proposed aiming to provide distributed ML, which are either centralized or decentralized: 

\point{Centralized}
In~\cite{7447103}, authors exploit the fact that the optimization algorithms used in modern deep learning can be parallelized and executed asynchronously. They propose distributed Selective Stochastic Gradient Descent (SSG) for enabling multiple parties to jointly learn an accurate neural-network model for a given objective without sharing their input datasets. In~\cite{10.5555/2997046.2997105}, authors propose an algorithm where each party trains a local classifier from its own data, and a third party aggregates these classifiers in a privacy-preserving manner using a cryptographic scheme and global differential privacy (DP). In~\cite{pmlr-v22-rajkumar12}, authors present a differentially private algorithm that  optimizes the overall multiparty objective rather than combining classifiers learned from optimizing local objectives. They achieve a weaker form of DP than~\cite{10.5555/2997046.2997105}, but provide improved generalization guarantees that do not depend on the number of parties or the relative sizes of the individual data sets.  In~\cite{Wang_2020}, authors propose a privacy preserving solution for Distributed Machine Learning (DML), where privacy protection is achieved through data randomization at the users' side and a modified ADMM algorithm is designed to mitigate the effect of data perturbation. In~\cite{8456019}, authors design a framework that focuses on the parameters with large absolute gradients to reduce privacy budget consumption during a SGD training. Authors adopt a generalization of the Report-Noisy-Max DP algorithm to select these gradients and prove its privacy guarantee. 
 
In~\cite{9282830}, authors aim at decentralizing FL by using mobile agents coupled with the FedAvg algorithm. Bonawitz \cite{10.1145/3133956.3133982} designed a failure-robust protocol for secure (based on secure MPC) aggregation of high-dimensional data even if an arbitrarily chosen subset of users drop out at any time. In~\cite{8241854}, authors present an enhanced system that utilizes asynchronous stochastic gradient descent and (similar to \toolname) additively HE. Finally, in~\cite{10.1007/978-3-030-41579-2_40}, authors design two protocols for privacy-preserving linear regression, Multilayer perceptron, and Convolutional neural network. One resistant to semi-honest and one resistant to servers-only malicious adversaries. 

\point{Decentralized}
In~\cite{shayan2019biscotti}, authors propose a system that leverages blockchain and cryptographic primitives to coordinate a privacy-preserving ML process between nodes. Authors evaluate their approach and demonstrate that it is scalable, fault tolerant and robust when 30\% of adversaries are trying to poison the model. In~\cite{9252063}, authors present a novel Byzantine-resilient distributed algorithm to train personalized models across similar peers, and convey theoretically and empirically the effectiveness of their approach in terms of speed of convergence, as well as robustness to Byzantine behavior. In~\cite{guo2020towards}, authors propose an algorithm to provide a uniform Byzantine-resilient aggregation rule for decentralized learning systems. In~\cite{10.1007/978-3-030-23502-4_10}, authors present their approach driven by a novel Leader-Follower (LF) topology and a DP model. 

In~\cite{9502443}, authors propose a a peer-to-peer variant of Federated Optimization by introducing a simple n-out-of-n secret sharing schema and an algorithm to calculate average values in a peer-to-peer manner. In~\cite{pappas2021ipls}, authors propose IPLS, which is partially based on the IPFS.  Similarly, in~\cite{hu2021gfl}, authors propose GFL, a decentralized FL framework based on IPFS and blockchain that introduces the consistent hashing algorithm to improve communication performance, and proposes a novel ring decentralized FL algorithm to improve decentralized FL performance and bandwidth utilization. 

\point{Our Approach} 
To the best of our knowledge, \toolname is the first \pp approach jointly providing several critical features. First, it can work with infrastructureless setups, as training is performed via proximity and cross-device communications and does not require any bootstraping (contrary to systems relying on blockchain~\cite{shayan2019biscotti}). Second, it is resilient to user churn and it is tolerant to peer failures, thanks to its synergy management mechanisms. Finally, it preserves the confidentiality of the shared gradients, and eliminates information leakage and poisoning attacks, without affecting the model's utility (contrary to approaches relying only on Differential Privacy~\cite{10.5555/2997046.2997105}).

%% file: sections/3_threat.tex
\section{Threat model}
\label{sec:threat}

\point{Users}
We assume a set of non-colluding users (or peers) that aim to train a machine learning model based on their local privacy-sensitive data, while also considering other nearby users' models. These users are equipped with mobile devices that support proximity and cross-device communication~\cite{nearby,airdrop} (\ie over Bluetooth Low Energy, Wifi Direct). We assume these users to be highly mobile and frequently co-locate physically during their everyday life.

Every peer following our protocol may play different roles: the one of (1) a synergy initiator or the one of (2) a participant. Each one of them owns a set of training examples drawn from its local data distribution over a feature space $X$ and a label space $Y$, defining a personal supervised learning task. The size of the training samples may vary among peers (depending on their activity); peers may have limited or no data at all (\eg a device with small storage capacity or new peers joining the network). Therefore, they may have very different models in terms of accuracy. Any exchange with a nearby peer may improve or weaken the peer's model, depending on the quality of the received model.

\point{Adversaries}
Similar to related works~\cite{244032, 9502443}, \toolname is designed for the  two-party semi-honest setting, where only one of the parties (initiator or participants) is corrupted by adversaries. Specifically, we assume the presence of curious peers that try to learn information about the other peers' private inputs from the
messages it receives. Specifically a corrupted peer is capable of:
\begin{enumerate}[itemsep=0pt,topsep=0pt,leftmargin=0.5cm]
    \item eavesdropping communications (\ie \textit{information leakage}), thus aiming to extract interests and/or preferences from the participating peer's training data and model parameters (Data Reconstruction~\cite{DRA}).
    
    \item tampering the communications or impersonating initiator peers.
    
    \item launching \textit{poisoning attacks} by acting as Byzanting peers that participate in synergies (\eg by performing label flipping or shuffling) with the intention of causing interference and corrupting the shared model (see Section~\ref{ssec:byzantine} for more details).
\end{enumerate}
Following the work done in similar settings~\cite{NIPS2017_f4b9ec30, 9252063}, we assume that such Byzantine peers have unlimited resources and comprise between $10-30\%$ of the total peer network.

\point{Assumptions}
\label{sssec:assumptions}
A created peer-to-peer group (for convenience, we will call it \emph{synergy}) always includes more than 2 peers. To ensure that, no peer ever sends their gradients back to the previous peer. \toolname does not assume the existence of \textit{any infrastructure or Internet connection}.

\subsection{Building Blocks}
\point{Confidentiality}
\label{ssec:confidentiality}
\toolname uses an additive Homomorphic Encryption (HE) scheme to perform \emph{private federated aggregation}. Upon installation, \toolname utilizes the device's 
hardware-backed keystore to generate an asymmetric public-private key-pair $(pk, sk)$. The public key encryption scheme is defined by three functions: (1) the encrypt function, where given a public key $pk$ and a message $M$, it outputs a ciphertext, $C=Enc(pk, M)$, (2) the decrypt function, that given a ciphertext and a private key $sk$, it outputs a decrypted message, $P=Dec(sk, M)$. 

The digital signature scheme is defined by a signing function, where given a message $M$ and a secret key $sk$, outputs a signature on the message, $S=Sign(sk, M)$ and a signature verification function that given a public key $pk$ and a signature $S$, it outputs a verified signature $V=Verify(pk, S)$. 

The additive homomorphic property guarantees that the addition of two ciphertexts $C1=Enc(pk, M_1)$ and $C2=Enc(pk,M_2)$, encrypted under the same key, results in the addition of the encryption of its messages:
\begin{equation*}
    Enc(pk,M_1)+Enc(pk,M_2) = Enc(pk,M1+M2)
\end{equation*}
In our \toolname prototype we use the Paillier partial homomorphic encryption scheme over elliptic curves \cite{10.1007/3-540-48910-X_16}.

%% file: sections/4_design.tex
\section{System Design}
\label{sec:design}
\toolname is an SDK that enables mobile apps to perform peer-to-peer (p2p) ML training using Stochastic Gradient Descent (SGD). Such apps may include various ML-involving cases like image classification, personalized advertising etc. App providers distribute a basic model within their app package and by using \toolname let their users (peers) converge to an optimal  global model, without jeopardizing their privacy. The peers' training data are kept private at all time, within their fortified walls of their device. 
Peers, interested in improving the app's model, form 
\emph{synergies} (either by initiating one or participating in other peers' synergies). In these synergies, they share their local model's parameters in a confidential way (via using strong cryptographic primitives) ensuring that (1) there is no information leakage and (2) the utility of the data is preserved.

\toolname runs in the background and in parallel to the user's daily activities (\ie while they socialize or when they co-locate with other people). If the user has opted in for this collaborative learning task, the application will (1) listen for ongoing synergies in the close proximity (or initiate its own), (2) share privately its local model parameters, and hear back the (privately) aggregated model from the synergy's initiator.
The frequency of synergy participation/initiation depends on the use case and is configurable by the app provider.

Next, we present the protocol of our approach in more detail. For convenience, we assume an image classification problem, where users use an application that includes a basic model for automatically grouping photos into groups~\cite{nguyen2022federated, li2020federated}. Users initially label a small set of photos to help the app retrain and personalize its basic model. 

\subsection{\toolname Protocol Description}
\begin{figure}
    \centering
    \includegraphics[width=0.4\textwidth]{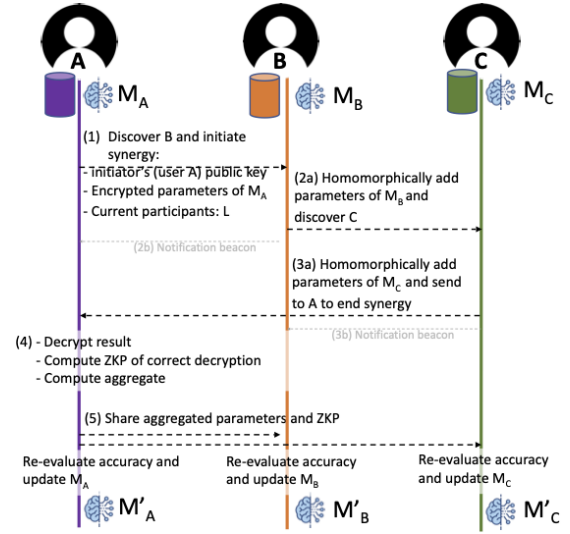}\vspace{-0.1cm}
    \caption{High-level overview of a 3-peers synergy in \toolname. The initiator creates a synergy where participants collaboratively perform private aggregation via HE.}
    \label{fig:protocol}\vspace{-0.6cm}
\end{figure}

Considering the above use case, we assume a user $A$ (\ie the synergy initiator) running an app for image classification that 
uses a local model $M$ trained on the user's personal data (user provided some initiate labels). 
In step 0, as seen in Figure~\ref{fig:protocol}, \toolname (1) utilizes the device's hardware-backed keystore to generate and store locally an asymmetric key-pair $(pk_A, sk_A)$, (2) extracts the parameters of model $M_A$
\begin{equation*}
    W_A: [W_{A1}, W_{A2}, W_{A3}, ..., W_{An}]
\end{equation*}
and (3) encrypts each one of them with the public key $pk_A$:
\begin{equation*}
    \begin{split}
    Enc(pk_A, {W_A}): [Enc(pk_A, W_{A1}), Enc(pk_A, W_{A2}), \\..., Enc(pk_A, W_{An})]
    \end{split}
\end{equation*}
In step 1, this user co-locates (\eg user arrives at their office) with $U$ other users of this service, who have their own personalized version of the model $M$ (in Table~\ref{tbl:notation} provides a summary of the notation). \toolname on the background starts advertising its presence and discovering other nearby peers running the same app. As soon as peer $B$ is discovered, peer $A$ creates a synergy by transmitting a (signed with its private key $sk_A$) message $\epsilon$ that includes its public key $pk_A$, the encrypted parameters of its model $Enc(pk_A, {W_A})$, an integer $S \leq U$ denoting the maximum remaining synergy participants, a timestamp $T$, and a list of the synergy participants $L: \{A_{addr}\}$:
\begin{equation*}
\epsilon = Sign(sk_A, [pk_A, Enc(pk_A, W_A), S, L, T])
\end{equation*}

In step 2, peer $B$ receives $\epsilon$ and after successfully verifying it, homomorphically adds the parameters $W_B$ of its own version of the model $M_B$ by using the public key of $A$:
\begin{equation*}
    \begin{split}
    [Enc(pk_A, W_{A1})+Enc(pk_A, W_{B1}), \\ Enc(pk_A, W_{A2})+Enc(pk_A, W_{B2}), ...,\\ Enc(pk_A,W_{An})+Enc(pk_A, W_{Bn})]
     \end{split}
\end{equation*}
which via its additive property is equal to:
\begin{equation*}
    \begin{split}
    Enc(pk_A, W_{A}+W_{B}) =  [Enc(pk_A, W_{A1}+W_{B1}), \\ Enc(pk_A,W_{A2} + W_{B2}),..., Enc(pk_A, W_{An}+W_{Bn})]
    \end{split}
\end{equation*}
Next, peer $B$ decreases $S$ by one, updates $L$ with its address: $L': \{A_{addr}, B_{addr}\}$, and discovers the next peer $C$ to forward the updated (signed with its own private key $sk_B$) $\epsilon'$:
\begin{equation*}
\epsilon' = Sign(sk_B,([pk_A, Enc(pk_A, W_A+W_B), S-1, L', T'])
\end{equation*}
As soon as $\epsilon'$ is received by peer $C$, $B$ sends a beacon to notify peer $A$.
Similarly, in step 3, the peer $C$ will add its model parameters and create $\epsilon''$ with $L'': \{A_{addr}, B_{addr}, C_{addr}\}$. If by decreasing $S$ by one this becomes zero, it means that $C$ was the last node to add its parameters, so it will not need to discover the next node to forward $\epsilon''$ but instead will send it to the first node in $L''$ (\ie $A_{addr}$). This time the signed $\epsilon''$ is simply:  
\begin{equation*}
\epsilon'' = Sign(sk_C, ([Enc(pk_A, W_A+W_B+W_C), L'', T''])
\end{equation*}
As soon as $\epsilon''$ is received by $A$, peer $C$ sends a beacon to notify $B$.
In step 4, peer $A$ receives $\epsilon''$, learns the number $N$ of participants from $L''$ and by using its private key $sk_A$, it decrypts $Res = Dec(sk_A(Enc(pk_A, W_A+W_B+W_C)) = \sum _{i=0}^{N-1} W_i)$ to compute in essence the aggregate:
\begin{equation*}
Aggr.Res = \frac{Res}{N} = \frac{Dec(sk_A(Enc(pk_A, \sum _{i=0}^{N-1} W_i))}{N}
\end{equation*}
where $N<=S$, and create a proof $\Pi_{Res}$ of correct decryption. As a final step (step 5), peer $A$ forwards across all participants in $L''$ the following:
\begin{equation*}
    [Aggr.Res, \Pi_{Res}, T]
\end{equation*}
This way, the rest of the synergy participants can learn the aggregated model parameters (following the concept of \emph{Federated Averaging}~\cite{mcmahan2017communication}) and verify the integrity of peer $A$'s decryption. Then every participant can update its local model with the aggregated parameters and re-evaluate its accuracy on the local data. If the accuracy of the updated model is lower than before the updates are discarded.
For simplicity, in this example, we assume only 1 synergy of just 3 participants, in reality, a user may participate in various synergies of various sizes in parallel (either as a synergy initiator or a participant).

\subsection{Key Properties}
After describing the overview of our protocol, we present the properties of our \pp system.

\point{User Churn} The dynamics of peer participation, or \emph{churn}, constitute an inherent property of \pp systems, critical to its design and evaluation. \toolname is designed to be dynamic thus allowing users to join and exit synergies without impeding the seamless operation of the synergy. \toolname assumes mobile users; this means that it is highly likely for users to frequently move in/out of range during an ongoing synergy.

\noindent\underline{Arriving users:} In \toolname, there is no initial, static selection of the synergy participants. Instead, every peer chooses the next participant, which means that any nearby peer may be selected to join the synergy at any time. This property allows \toolname synergies to be formed on-the-fly and select a different topology in case of a failed (or out of range) participant. 

\noindent\underline{Departing users:} A participating peer may depart from a synergy (connectivity loss, application/device fail) without forwarding its updated $\epsilon$ causing this way an interruption in the synergy's expected operation. For instance, in the above example, peer $B$ may discover peer $C$, forward $\epsilon'$ and then peer $C$ may go out of range. 

To avoid such cases of incomplete synergies, as we described above, as soon as peers successfully forward their updated $\epsilon$ to their neighbour, they notify the previous peer with a \emph{notification beacon}. In \toolname there is a specified time window $t_p$ for a peer to compute and forward its updated $\epsilon$ to the next peer. As a result, after $t_p$ time without receiving a notification beacon, the previous peer knows that its neighbour dropped the synergy. Then, as a fail-over, the previous peer needs to select another neighbour to forward $\epsilon'$. After $R$ unsuccessful retries, and if the peer was not only the second participant in the synergy (synergy size must always be $>2$ as defined in Section~\ref{sec:threat}), it will sent $\epsilon'$ to the synergy's initiator so the synergy can end. If a synergy initiator does not receive any response from a peer after time $t=t_p\times (S-1)$, the synergy is considered a failure.

\begin{table}[t]
    \centering
    \scriptsize
    \begin{tabular}{ll}
    \toprule
        \bf Notation & \bf Explanation \\ \midrule
            $U$	& Total number of users \\
            $S$	& Max number of synergy participants requested \\
            $pk_i$ &	Public key of peer i\\
            $sk_i$ &	Private (secret) key of peer i\\
            $W_{ij}$ &	j$^{th}$ weight of peer i\\
            $i_{addr}$ &	The network address of the peer i\\
            $L$ &	A list that includes the network addresses of  the participating peers\\
            $\epsilon$ &	Transmitted message from one peer to an other\\
            $Res$	& Decrypted result in plaintext\\
            $\Pi_{Res}$ &	Proof of correct decryption for Res \\
         \bottomrule
    \end{tabular}
    \caption{Summary of Notation}\vspace{-0.8cm}
    \label{tbl:notation}
\end{table}

\point{Confidentiality of user model}
\toolname protects against inference attacks by ensuring that no peer can learn the model parameters of \emph{any} individual synergy participant. Instead, only the aggregated averages are shared, meaning that the more participants are in a synergy the harder it becomes for the curious eavesdropper to infer the model parameters of the participants.
Contrary to related works that utilize obfuscation-based techniques (\eg differential privacy),
\toolname employs strong cryptographic primitives for the private aggregation of the user models' parameters. This means that it does not incur any 
utility loss in the aggregated data, which allows the model propagation algorithm to converge faster (see Section ~\ref{ssec:encryption_loss}).

\point{Infrastructureless setup and Fault Tolerance}
Contrary to centralized, hybrid or decentralized distributed and Federated learning approaches (see Table~\ref{tab:related}), \toolname does not rely on \emph{any form of infrastructure} to successfully carry out the collaborative learning process. In our approach, peers can privately share their models' parameters without requiring a server or even Internet connection, by leveraging BLE and Wifi Direct protocols (as provided by Android Nearby and Apple AirDrop). This means that \toolname is fault-tolerant, since a potential peer failure would affect only the synergy. Specifically, (1) if the failed peer is the initiator, the ongoing synergy would collapse before the aggregated parameters are shared, (2) if the failed peer is a participant, then the synergy will continue unaffected; even if the participant fails while handling the synergy message, the previous participant will detect the missing notification beacon and will redirect the synergy message to another neighbour.

%% file: sections/5_simulations.tex
  \section{Simulation-based Evaluation}
\label{sec:simulation}

To investigate the properties of the \toolname, we implement a peer-to-peer simulation network that enables end-to-end communication between peers of the network. Our simulations aim to study the effect of the following factors: (1) Effect of synergy size, (2) Effect of data imbalance (ID vs. non-IID), (3) Loss due to encryption, and (4) Effect of Byzantine peers. Furthermore, we compare and contrast our approach with two baseline models: (1) a model trained in a centralised manner, and (2), a centralised FL approach of collaborative ML training. In what follows, we discuss the implementation environment, the evaluation metrics and obtained results.

\subsection{Simulation Environment}
\label{ssec:implementation_environment}

\point{Datasets}
For our evaluation, we use three large-scale datasets, \texttt{CIFAR-10}~\cite{krizhevsky2009learning}, \texttt{Avito}~\cite{avito} and \texttt{IMDB}~\cite{maas2011}, suitable for the on-device ML setting we consider. The \texttt{CIFAR-10} is a standard choice among many FL works~\cite{konevcny2016federated, zhao2018federated, Nasr19}, and we include it in our analysis for consistency and comparability purposes. It contains 60,000 $32\times32$ colour images, across 10 classes, with 6,000 images per class. The \texttt{Avito} dataset, on the other hand, is a realistic dataset of over 4M users that has been used in competitions for a click-through rate prediction task. Contrary to other popular options used for FL research (e.g., Shakespeare ~\cite{abs-1812-01097}, Federated EMNIST~\cite{abs-1812-01097}), \texttt{Avito} has several desirable properties: 1) the data follow a realistic (Non-IID) distribution across the users, which translates easily to a p2p network where each peer contributes a different number of samples, and 2) it is characterised by a skewed class distribution, which makes it particularly challenging for training ML models, especially on partial local data. Finally, the \texttt{IMDB} sentiment classification dataset consists of 50,000 movie reviews that are labeled as either positive or negative. This dataset is appropriate for natural language processing tasks that are not addressed by the previous two datasets.

\point{Experimental settings}
In our simulations, we consider a set of peers to train an ML model using their local data, while considering other peers' models in the network. More formally put, each peer $i$ owns a set of $k_{i}$ training samples $S_{i} = \{x^{j}_{i}, y^{j}_{i}\}^{k_{i}}$ drawn from the peer's local distribution ${\mu_{i}}$ over a feature space $X$ and a label space $Y$, defining a personal supervised learning task. We consider both IID and non-IID settings. Then, given a convex loss function $\ell:\mathbb{R}^{p}\times X \times Y$, the objective of peer $i$ is to learn a model $\theta_{i} \in \mathbb{R}^{p}$ with a small expected loss $\mathbb{E}_{x_{i}, y_{i}} \backsim \mu_{i}\ell(\theta_{i}, x_{i}, y_{i})$. 
Each peer $i$ learns a local model by minimizing its local loss over $S_{i}$:
\begin{equation*}
\theta^{local}_{i} \in \underset{\theta \in \mathbb{R}^{p}}{\mathrm{argmin}} \mathcal{L}_{i}(\theta) = \sum_{1}^{k_{i}} \ell(\theta_{i}, x_{i}, y_{i})
\end{equation*}
We note that, for the non-IID setting, the size of the local training data is variable across peers i.e., some peers may have available very few or no data at all (\eg a mobile device with limited storage). As a result, the local models will differ in terms of performance and any exchange of model parameters may degrade or improve a local model, depending on the utility of the received weights.

\subsubsection{Machine learning tasks}
\label{sssec:ml_tasks}

We evaluate \toolname on three ML tasks. For our first task, we run simulations with \texttt{CIFAR-10} and cast it as a multi-class classification problem. We consider a network of 1,000 peers (devices) with 50 samples each, where the probability of one sample to belong to one class is the same for every peer in the IID setting, while for the non-IID setting we follow a strategy where each user only has samples from 6 non-overlapping classes\footnote{This number varies across studies, e.g., in \cite{hu2021gfl} the authors use 2 classes.}. For testing purposes, we hold out 10,000 samples. Furthermore, for consistency with prior FL research with \texttt{CIFAR-10}, we apply the following parameter values: \texttt{batch\_size}=16, \texttt{epochs}=50, minimum round threshold for peers \texttt{mrt}=50, maximum ratio of peers (that must meet \texttt{mrt}) \texttt{mrr}=0.5, and use an SGD optimizer with \texttt{L1}=0.001. We perform our simulations by sampling from the standard uniform distribution, assuming a constant probability $P_{i} = \frac{1}{N}$ of selecting peer $i$ (where $N$ is the total number of peers in the network) to participate in a synergy and, in addition, we derive random samples from a power-law probability distribution
    $f(x,a)=ax^{a-1}$, 
where $0 \leq x \leq 1$ and $a=2$. For our performance analysis, we report the metrics of loss, Accuracy and AUC as a function of the number of communication rounds (i.e., iterations) performed to reach convergence.

\begin{figure*}[!t]
    \captionsetup[subfloat]
    {}
    \centering
    \subfloat[]{
    \label{figure:sz_cifar_iid_a}
    \includegraphics[clip=true, trim=0 0 0 0, width=0.3\textwidth]{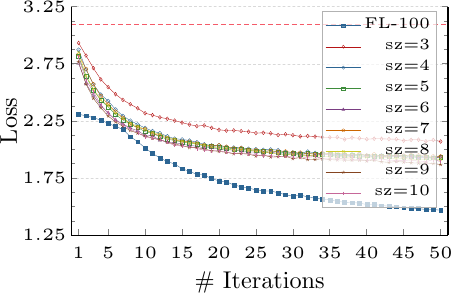}
    }
    \hspace{0.05mm}
    \subfloat[]{%
    \label{figure:sz_cifar_iid_b}
    \includegraphics[clip=true, trim=0 0 0 0, width=0.3\textwidth]{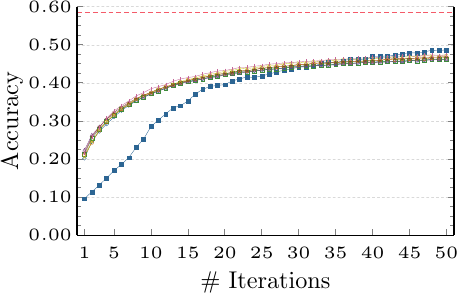}
    }
    \hspace{0.05mm}
    \subfloat[]{%
    \label{figure:sz_cifar_iid_c}
    \includegraphics[clip=true, trim=0 0 0 0, width=0.3\textwidth]{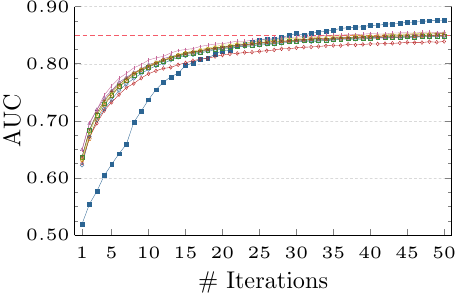}
    }
    \vspace{-0.75em}
    \caption{Effect of synergy size on Loss (left), Accuracy (middle) and AUC (right) on IID data for \texttt{CIFAR-10} (\texttt{mrt}=50, \texttt{mrr}=50). Red dotted lines denote baseline performance. 
    }\vspace{-0.4cm}
    \label{figure:sz_cifar_iid}
\end{figure*}

\begin{figure*}[!t]
    \captionsetup[subfloat]
    {}
    \centering
    \subfloat[]{
    \label{figure:sz_cifar_non_iid_a}
    \includegraphics[clip=true, trim=0 0 0 0, width=0.3\textwidth]{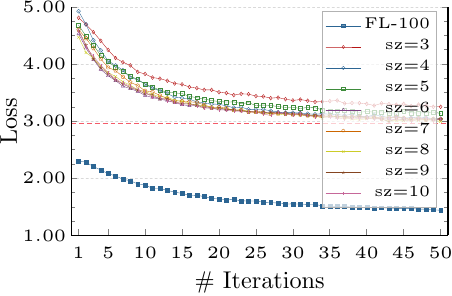}
    }
    \hspace{0.05mm}
    \subfloat[]{%
    \label{figure:sz_cifar_non_iid_b}
    \includegraphics[clip=true, trim=0 0 0 0, width=0.3\textwidth]{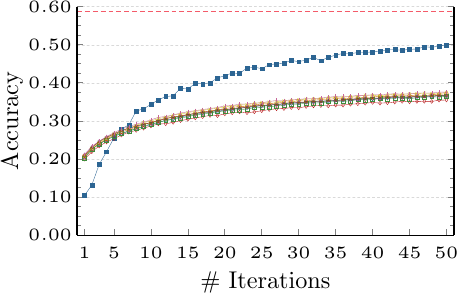}
    }
    \hspace{0.05mm}
    \subfloat[]{%
    \label{figure:sz_cifar_non_iid_c}
    \includegraphics[clip=true, trim=0 0 0 0, width=0.3\textwidth]{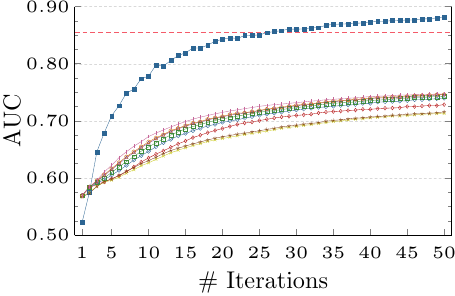}
    }
    \vspace{-0.75em}
    \caption{Effect of synergy size on Loss (left), Accuracy (middle) and AUC (right) on non-IID data for \texttt{CIFAR-10} (\texttt{mrt}=50, \texttt{mrr}=50). Red dotted lines denote baseline performance. 
    }\vspace{-0.4cm}
    \label{figure:sz_cifar_non_iid}
\end{figure*}

For our second task, we run simulations with \texttt{Avito} and cast it as a logistic regression problem. We apply a series of pre-processing steps on the original dataset (\eg joining of tables, removing features with zero-variance, applying one-hot-encoding to categorical variables) that produces a final dataset with 127 numeric and categorical features. We then derive a random sample of 1,000 users for the FL baseline and a random sample of 200 users ($\sim$250,000 samples) for our peer training network, while retaining 50 users for testing ($\sim$65,000 samples). \texttt{Avito} is a non-IID dataset (each user contributes a variable number of data points). Given its highly imbalanced class distribution (class 0: $99.39\%$
), we report only loss and AUC~\cite{5128907}, and omit Accuracy due to its sensitivity to imbalanced data. We consider the following configuration: 
\texttt{batch\_size}=128, \texttt{epochs}=3, \texttt{mrt}= 50, \texttt{mrr}=0.5, SGD optimizer with \texttt{L1}=0.001 and \texttt{L2}=0.001.

For our third task, we run simulations with \texttt{IMDB} and cast it as a natural language processing problem. Similarly to \texttt{CIFAR=10}, we situate our analysis in a non-IID setting. We consider a randomly sampled network of 1,000 peers (devices) for the FL baseline and one with 200 users for our peer training network, while reserving the pre-determined test split of 25,000 reviews for testing purposes. The \texttt{IMDB} movie review sentiment classification dataset consists of reviews that have been pre-processed, and every review is encoded as a sequence of term indices. The terms within the reviews are indexed by their frequency count within the dataset, e.g., index ``2'' encodes the second most frequent term in the data. In our analysis, we consider only the top 10,000 most common terms, while setting the maximum sequence length to 500 (longer sequences are truncated). We consider the following parameter configuration: 
\texttt{batch\_size}=32, \texttt{epochs}=3, \texttt{mrt}= 50, \texttt{mrr}=0.5, SGD optimizer with \texttt{L1}=0.001 and \texttt{L2}=0.001.

Finally, all simulations were carried out on a server running Ubuntu 18.04.2 LTS, with 2 $\times$ Intel(R) Xeon(R) CPU E5-2660 v4 at 2.0 GHz, with 56 cores, 512Gb RAM and 7 $\times$ GeForce GTX 2080 TI (11 Gb). The source code\footnote{Link to open source repository omitted to preserve anonymity} was written in Python v3.8 and uses the Keras framework of TensorFlow v2.6.0.

\subsubsection{Baselines and shared models}
\label{sssec:baselines_and_peer_models}

We selected two popular approaches as baselines: (1) Classical ML baseline: a classical centralised ML framework where the variants of the shared models use all the peer training data, and (2) FL baseline: a centralised federated learning~\cite{mcmahan2017communication}, where a central server aggregates the parameters from the locally (at the user's device) trained models before sending them back to the user's device. The Classical ML baseline variant is trained using the data from all the peer network and is evaluated against the held-out dataset. Our FL baseline is trained for 50 (FL) rounds and, during each round, we sample 100 random participants following a power law distribution to simulate real-world conditions (where some users participate more often than others) and keep our experiments consistent with the \toolname ones.

As a shared model for the multi-class classification task with \texttt{CIFAR-10} we use \texttt{MobileNetv2}~\cite{sandler2018mobilenetv2}, which has been pre-trained on the ImageNet~\cite{5206848} dataset with image size $224 \times 224$. To personalise the model, we augment it with two dense layers, one with $62,720$ neurons and an additional with $32$ neurons, followed by a softmax activation (SGD optimizer with \texttt{L1}=0.001), amounting to a total of $2,007,402$ trainable parameters. For the Classical ML baseline model we use the following parameters: \texttt{epochs}=300, \texttt{patience}=10, and \texttt{min-delta}=0.001. For the FL baseline we use \texttt{epochs}=5. In both approaches, we use \texttt{batch\_size}=16 and an SGD optimizer with \texttt{L1}=0.001 and \texttt{L2}=0.0001.

Since logistic regression models can be viewed as special case of neural networks (i.e, a single layer model, without any hidden layers), we implement the shared model for the logistic regression task with \texttt{Avito} as a single-layer densely connected network with $127$ neurons, followed by a softmax activation (SGD optimizer with \texttt{L1}=0.001 and \texttt{L2}=0.001), with a total of $127$ trainable parameters. For the Classical ML baseline model we use the following parameters: \texttt{epochs}=300, \texttt{patience}=10, and \texttt{min-delta}=0.001. For the FL baseline, we use \texttt{epochs}=5. In both approaches, we use \texttt{batch\_size}=128 and an SGD optimizer with \texttt{L1}=0.001 and \texttt{L2}=0.0001.
For the natural language processing task with \texttt{IMDB}, we implement the shared model as a multi-layer NN consisting of an embedding layer (so that our NN becomes more capable of extracting the meaning of terms), stacked on top of a Bidirectional Long Short-Term Memory (BLSTM) layers, and two dense layer with interleaving dropout layers (for reducing overfitting) and 1D global max pooling (for extracting better patterns and generating spatial hierarchy), followed by a softmax activation (SGD optimizer with \texttt{L1}=0.001 and \texttt{L2}=0.001), amounting to a total of $1,547,314$ trainable parameters. For the Classical ML baseline model we use the following parameters: \texttt{epochs}=300, \texttt{patience}=10, and \texttt{min-delta}=0.001. For the FL baseline, we use \texttt{epochs}=5. In both approaches, we use \texttt{batch\_size}=32 and an SGD optimizer with \texttt{L1}=0.001 and \texttt{L2}=0.0001.

\begin{figure}[!t]
    \captionsetup[subfloat]
    {}
    \centering
    \subfloat[Loss]{
    \label{figure:sz_avito_non_iid_a}
    \includegraphics[clip=true, trim=0 0 0 0, width=0.24\textwidth]{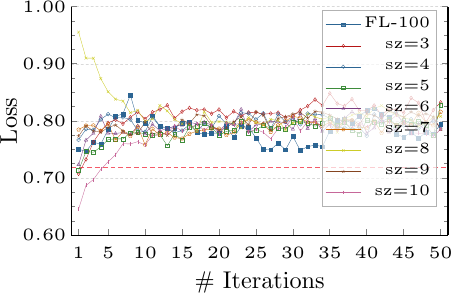}
    }
    \subfloat[AUC]{%
    \label{figure:sz_avito_non_iid_c}
    \includegraphics[clip=true, trim=0 0 0 0, width=0.23\textwidth]{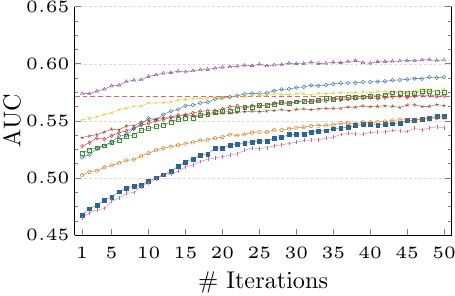}
    }
    \vspace{-0.75em}
    \caption{Effect of synergy size on Loss and AUC on non-IID data for \texttt{Avito} (\texttt{sz}=3-10, \texttt{mrt}=50, \texttt{mrr}=50). Red dotted lines denote baseline performance. 
    }
    \label{figure:sz_avito_non_iid}
    \vspace{-0.3cm}
\end{figure}

\begin{figure*}[!t]
    \captionsetup[subfloat]
    {}
    \centering
    \subfloat[Loss]{
    \label{figure:sz_imdb_non_iid_a}
    \includegraphics[clip=true, trim=0 0 0 0, width=0.3\textwidth]{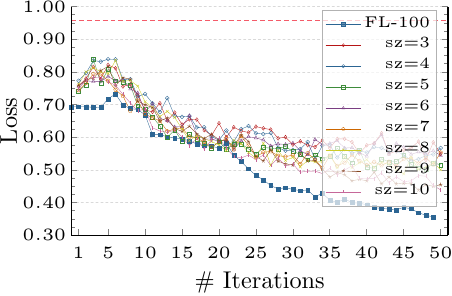}
    }
    \hspace{0.05mm}
    \subfloat[Accuracy]{%
    \label{figure:sz_imdb_non_iid_b}
    \includegraphics[clip=true, trim=0 0 0 0, width=0.3\textwidth]{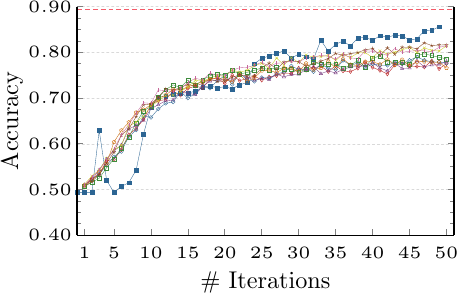}
    }
    \hspace{0.05mm}
    \subfloat[AUC]{%
    \label{figure:sz_imdb_non_iid_c}
    \includegraphics[clip=true, trim=0 0 0 0, width=0.3\textwidth]{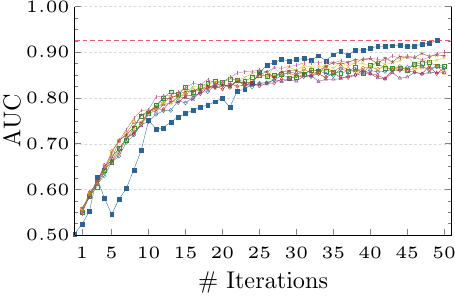}
    }
    \vspace{-0.75em}
    \caption{Effect of synergy size on Loss, Accuracy and AUC on non-IID data for \texttt{IMDB} (\texttt{sz}=3-10, \texttt{mrt}=50, \texttt{mrr}=50). Red dotted lines denote baseline performance. 
    }\vspace{-0.4cm}
    \label{figure:sz_imdb_non_iid}
\end{figure*}

\begin{figure*}[!t]
    \captionsetup[subfloat]
    {}
    \centering
    \subfloat[Loss]{
    \label{figure:crypto_imdb_non_iid_a}
    \includegraphics[clip=true, trim=0 0 0 0, width=0.3\textwidth]{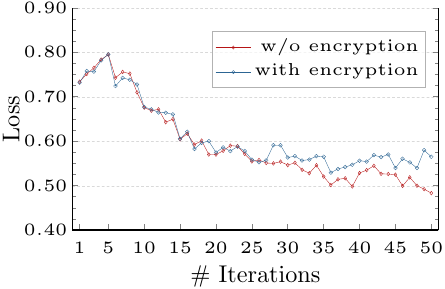}
    }
    \hspace{0.05mm}
    \subfloat[Accuracy]{%
    \label{figure:crypto_imdb_non_iid_b}
    \includegraphics[clip=true, trim=0 0 0 0, width=0.3\textwidth]{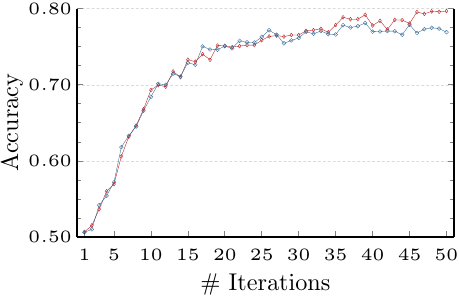}
    }
    \hspace{0.05mm}
    \subfloat[AUC]{%
    \label{figure:crypto_imdb_non_iid_c}
    \includegraphics[clip=true, trim=0 0 0 0, width=0.3\textwidth]{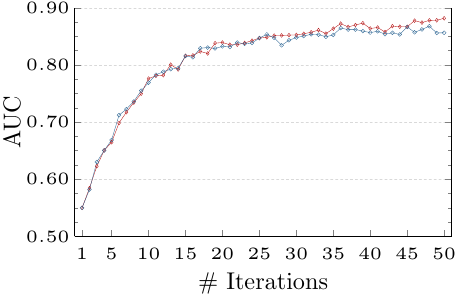}
    }
    \vspace{-0.75em}
    \caption{Effect of encryption on Loss, Accuracy, AUC on non-IID data for \texttt{IMDB} (\texttt{sz}=3-10, \texttt{mrt}=50, \texttt{mrr}=50). 
    }\vspace{-0.2cm}
    \label{figure:crypto_imdb_non_iid}
\end{figure*}

\subsection{Effect of Synergy Size $\times$ Data Imbalance}
\label{ssec:synergy_size_data_imbalance}

We begin by studying the effect of synergy size on the model training convergence behaviour. We initially consider fixed synergy sizes \texttt{sz} $\in [3-10]$ with our IID data. Figure \ref{figure:sz_cifar_iid} shows the results of the simulation-based evaluation for \texttt{CIFAR-10}. 
In Figure~\ref{figure:sz_cifar_iid_a}, we plot the average loss as a function of the number of rounds peers have participated in (for each data point we consider only those peers that have participated in \emph{at least} $N$ iterations). 
Here, we note that the loss of the \toolname model is reduced by approximately $20\%$ within the first five iterations. 
Figure~\ref{figure:sz_cifar_iid_b} compares the average Accuracy of the baseline models against the \toolname model. 
Considering that a multi-class classification problem such as \texttt{CIFAR-10} is particularly challenging, the Classical ML baseline achieves a reasonable Accuracy of $0.58$, while the \toolname and FL baseline models converge around the value of $0.47$ ($\lessapprox14\%$). When considering the average AUC (\ref{figure:sz_cifar_iid_c}), the \toolname model matches and, eventually, slightly exceeds the baseline performance ($0.84$) after the $30^{th}$ iteration. 

Next, we examine the effect of synergy size on the \toolname model training convergence behaviour for non-IID data. Figure~\ref{figure:sz_cifar_non_iid} shows the results of the simulation-based evaluation for \texttt{CIFAR-10}. Figure~\ref{figure:sz_cifar_non_iid_a} shows the average loss for various synergy sizes. Compared to the IID variant of our simulations, the staring loss of the \toolname model is about $65\%$ higher in the initial rounds, although it approximates the baseline loss after the $40^{th}$ iteration
. Also, Figures~\ref{figure:sz_cifar_non_iid_b} and \ref{figure:sz_cifar_non_iid_c} show a positive trend in the performance of the \toolname model w.r.t. to average Accuracy and AUC. The performance gap with the baseline model is reduced to $52\%$ and $13\%$ respectively after 40 iterations. As for the FL baseline, its performance is very close to, or better, than that of the Classical ML baseline.

In Figure~\ref{figure:sz_avito_non_iid} we show the results for \texttt{Avito} for the non-IID data. Due to the imbalanced class distribution of \texttt{Avito}, we plot only the average loss and AUC (in Figures~\ref{figure:sz_avito_non_iid_a} and \ref{figure:sz_avito_non_iid_c} respectively). Regarding loss, we observe that medium-to-large synergies (\texttt{sz} $\geq 6$) appear to converge faster and reach a loss that is only $9\%$ higher, compared to the baseline loss, and comparable to the loss of the FL baseline. With respect to AUC, the \toolname model outperforms the baseline model within 20 iterations, even for small synergy sizes (\texttt{sz} $\leq 6$), and by the $25^{th}$ iteration it converges to $0.61$, which is $9\%$ better than the AUC achieved by the FL baseline.

Last, in Figure~\ref{figure:sz_imdb_non_iid} we see the results for \texttt{IMDB} for the non-IID data. In terms of the loss metric (Figure~\ref{figure:sz_imdb_non_iid_a}), both the \toolname and the FL baseline follow a downward trend, eventually reaching losses superior ($> 63.6\%$ improvement) to that of the Classical ML baseline. With respect to average Accuracy (Figure~\ref{figure:sz_imdb_non_iid_b}), both \toolname and FL Baseline models reach competitive performances by the $50^{th}$ round, with the \toolname exhibiting a smoother and more stable convergence. As for average AUC (Figure~\ref{figure:sz_imdb_non_iid_c}), we note a very similar behaviour as before, with both \toolname and FL Baseline models competing toe to toe and, finally, approximating the Classical ML baseline performance by the $50^{th}$ round. We note that these initial findings demonstrate that the final global model, trained under \toolname, can achieve a competitive performance (in some cases even better than that of a centralised model). \toolname can clearly support distributed ML tasks, especially considering that it brings certain desirable privacy and security properties (Section~\ref{ssec:confidentiality}) that the centralised counterparts do not offer. Moreover, although in distributed optimization the local data of the peers are assumed to be IID, in a multi-party ML setting federated learning assumes that peers possess training data that are not identically and independently distributed (non-IID) across the peer network. Therefore, hereinafter, we report our findings only for the non-IID setting. In addition, for each synergy we sample our peers and synergy sizes (\texttt{sz}=3-10) from a power-law distribution to simulate more realistic conditions where some peers are expected to participate in synergies more often.

\subsection{Loss Due to Encryption}
\label{ssec:encryption_loss}
As discussed in Section~\ref{ssec:confidentiality}, we use additive homomorphic encryption to guarantee the privacy of the shared model parameters. Homomorphic encryption by definition deals with messages that are in integer-format only. Therefore, in our prototype, before encrypting the model parameters (floating-point numbers), we had fist to convert them to long integers (by multiplying each of them with $10^{10}$)\footnote{Similarly after decryption we divide by the same number}. However, other similar ML approaches such as quantization-aware training~\cite{quant} are also compatible with our technique. 

To assess the potential precision loss, we report the effect of encryption on our \toolname model. Figure~\ref{figure:crypto_imdb_non_iid} shows the loss, Accuracy and AUC for the \texttt{IMDB} dataset, with and without homomorphic encryption (the figures for the remaining datasets show very little variance and, therefore, are ommitted). For the most part, we don't observe any notable impact; only for \texttt{IMDB} we note a small degradation in model's performance in the range of $3-4\%$ (w.r.t. Accuracy and AUC). We attribute this loss in utility to the type of model architecture (RNN vs. CNN vs. NN), and not to its complexity or depth, since all models share a very large number of parameters (\eg $\geq$ 10K).

\subsection{Effect of Poisoning Attacks}
\label{ssec:byzantine}
In this section, we evaluate the performance of \toolname when we subject it to different types of attacks initiated by Byzantine peers, as done in previous works~\cite{9252063,shayan2019biscotti}. We investigate different parameter settings to study the training behaviour of the shared model and evaluate how well it performs under attack from 10, 20 and 30\% of Byzantine peers. We consider that every Byzantine peer aims to compromise the models of its neighbours by communicating random models in each communication step. Specifically, we demonstrate the impact of the following, relevant to our work, label poisoning attacks:

\input{figs/figureWall1}

\point{Label flipping attack (fixed classes)} where a certain class in the training data for the Byzantine peers is mislabeled, causing the model to learn to miss-classify it~\cite{10.1145/2046684.2046692,10.5555/3007337.3007488}. In our setup, this attack would consistently flip classes 0 (``airplane'' concept) and 5 (``dog'' concept) in \texttt{CIFAR-10}, and classes 0 and 1 in \texttt{Avito} and \texttt{IMDB}. 

\point{Label flipping attack (random classes)} where a random pair of classes in the training data for the Byzantine peers is mislabeled. Arguably, this a somewhat less effective variant of the label flipping attack with fixed classes, since it corrupts a random selection of class concepts every time (only applicable to \texttt{CIFAR-10}).

\point{Label shuffling attack (all classes)} where we shuffle all available labels in the training data for the Byzantine peers. This poisoning attack causes the most extensive damage to the underlying classification system of the shared model and forces adaptation to false data (only applicable to \texttt{CIFAR-10}).

\point{Noisy peers} similar to~\cite{NIPS2017_f4b9ec30}, adversarial peers will share model parameters that differ significantly to the ones from honest peers. Here, we assume that the Byzantine peers substitute their weights with new weights, randomly sampled from a similar distribution. More specifically, let $\theta$ denote the weight matrix consisting of the model weights $\theta^{j}_{i}$, where $j$ denotes the layer and $i$ denotes the unit in layer $j$. In a more general setting, $\theta^{j}$ can be viewed as the collection of all parameters of the \textit{j}$^{th}$ layer. At each iteration, we compute a kernel density estimate for the layer\footnote{For \texttt{CIFAR-10} we consider only the weights of the last two dense layers, while for \texttt{Avito} the single dense layer.} we are interested in, using a one-dimensional Gaussian kernel
\begin{equation*}
    G(x;\sigma) = \frac{1}{\sqrt{2\pi}\sigma}e^{\frac{-x^2}{2\sigma^{2}}}
\end{equation*}
where $\sigma$ is the standard deviation of the distribution that controls the width of the Gaussian kernel, to learn a non-parametric generative model of the weights $\theta^{j}$. More specifically, the density estimate at a weight $x$ within a set of weights $w_{i};i=1 \ldots N$ in $\theta^{j}$ is given by:
\begin{equation*}
    \rho_{G}(x) = \sum_{i=1}^{N}G(x - w; \sigma)
\end{equation*}
Using this generative model we efficiently draw new samples $\theta'^{j}$ at every iteration and perform the update: $\theta_{t+1}^{j} \leftarrow \theta_{t}'^{j}$. Such regular updates result in an adversarial concept drift that slows down the convergence of the shared model.

Figure \ref{figure:cifar_label-stealth-fixed} 
shows the effect of different poisoning attacks for \texttt{CIFAR-10} (we omit the figures for the label shuffling, noisy peers and label flipping for random classes attacks due to the similarity with the results shown here). Based on these preliminary findings, the main take-away message is that, irrespective of the type of attack, the effect on the models' performance (\toolname and FL Baseline) is negligible. This appears to be the case despite the lack of defense mechanisms or the high percentage of Byzantine peers in the network. 

\begin{figure*}[!t]
    \captionsetup[subfloat]
    {}
    \centering
    \subfloat[CPU utilization]{
    \label{figure:performance_a}
    \includegraphics[clip=true, trim=0 0 0 0, width=0.29\textwidth]{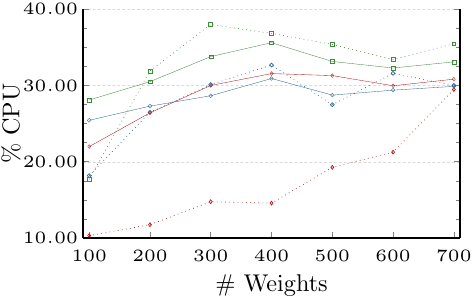}
    }
    \hspace{0.15mm}
    \subfloat[Power discharge]{%
    \label{figure:performance_b}
    \includegraphics[clip=true, trim=0 0 0 0, width=0.29\textwidth]{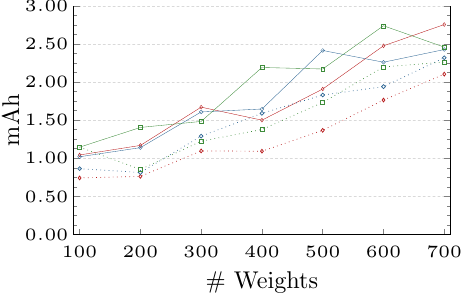}
    }
    \hspace{0.15mm}
    \subfloat[e2e synergy]{%
    \label{figure:performance_c}
    \includegraphics[clip=true, trim=0 0 0 0, width=0.28\textwidth]{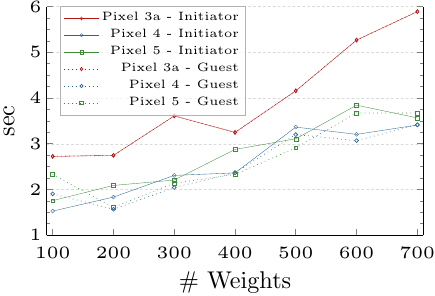}
    }
    \vspace{-0.75em}
    \caption{System performance in terms of (a) CPU utilization, (b) power discharge and (c) latency for e2e synergy, for the initiator and guest. 
    }
    \label{figure:performance}\vspace{-0.5cm}
\end{figure*}

A very different conclusion is drawn for the \texttt{Avito} dataset. We show in Figure~\ref{figure:avito_non_iid_label_flipping_a} only the results for the label flipping and in Figure~\ref{figure:avito_non_iid_label_flipping_c} the noisy peers  attacks (as discussed in Section~\ref{ssec:byzantine}). It is noteworthy that the impact of the noisy peers attack is more significant and primarily affects the performance of the \toolname model which, for both attacks, surpasses in performance the FL Baseline. In addition, we observe that the \% of Byzantine peers is the least influential factor, leaving the \toolname and the FL Baseline unaffected. In reference to the label flipping attack, it is our supposition that the absence of effect is, in some measure, owing to the imbalance in classes observed in practical datasets such as \texttt{Avito}. As the majority class typically dominates the label space of most peer-local data, this type of attack proves ineffective in undermining the classification system. In contrast, the noisy peers attack appears to be more effective in causing the shared model to converge towards an inferior solution.

Finally, Figures~\ref{figure:imdb_non_iid_label_flipping}-\ref{figure:imdb_non_iid_byzantine} illustrate the effect of the label flipping and the noisy peers attacks for the \texttt{IMDB} dataset. In this particular scenario, we find that the performance degradation is most pronounced in relation to the proportion of Byzantine peers in the network, across all models. However, the delta in terms of Accuracy and AUC lies within the range of $2-3\%$. Both \toolname and the FL Baseline maintain a competitive performance, with the \toolname being the ``early winner'' (outperforms the FL Baseline within the first 25-30 rounds) and the FL Baseline achieving slightly better performance from the $30^{th}$ round onward.

%% file: figs/figureWall1.tex
\begin{figure*}[!t]
\begin{minipage}[t]{0.49\textwidth}
    \captionsetup[subfloat]
    {}
    \centering
    \subfloat[Accuracy]{%
    \label{figure:cifar_label-stealth-fixed_b}
    \includegraphics[clip=true, trim=0 0 0 0, width=0.5\textwidth]{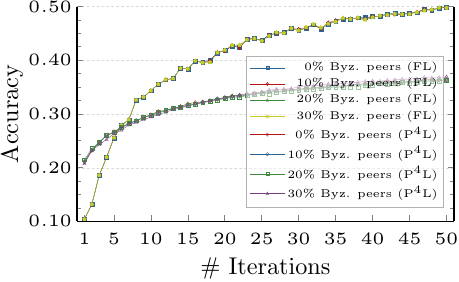}
    }
    \subfloat[AUC]{%
    \label{figure:cifar_label-stealth-fixed_c}
    \includegraphics[clip=true, trim=0 0 0 0, width=0.5\textwidth]{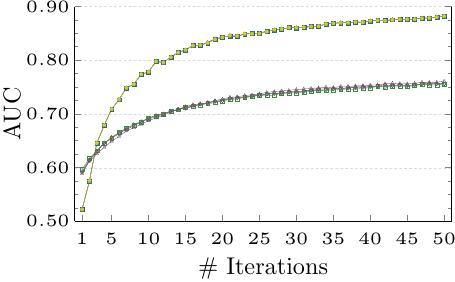}
    }
    \vspace{-0.75em}
    \caption{Effect of label flipping attack (fixed classes) on Accuracy and AUC for non-IID \texttt{CIFAR-10} (\texttt{sz}=3-10, \texttt{mrt}=50, \texttt{mrr}=50).  
    }
    \label{figure:cifar_label-stealth-fixed}
    \end{minipage}
\hfill
    \begin{minipage}[t]{0.49\textwidth}
    \captionsetup[subfloat]
    {}
    \centering
    \subfloat[label flipping attack]{
    \label{figure:avito_non_iid_label_flipping_a}
    \includegraphics[clip=true, trim=0 0 0 0, width=0.5\textwidth]{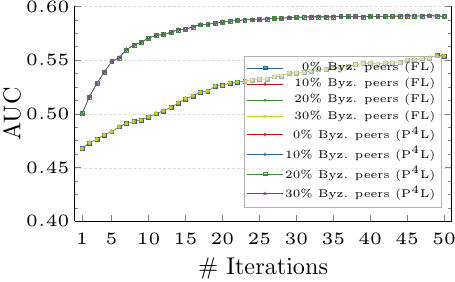}
    }
    \subfloat[noisy peers attack]{%
    \label{figure:avito_non_iid_label_flipping_c}
    \includegraphics[clip=true, trim=0 0 0 0, width=0.5\textwidth]{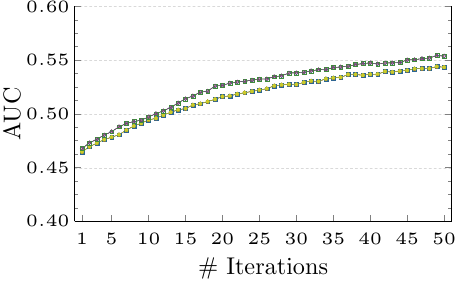}
    }
    \vspace{-0.75em}
    \caption{Effect on AUC of label flipping attack and noisy peers attack for \texttt{Avito} (\texttt{sz}=3-10, \texttt{mrt}=50, \texttt{mrr}=50). 
    }
    \label{figure:avito_non_iid_label_flipping}
    \vspace{-0.4cm}
\end{minipage}
\end{figure*}

\begin{figure*}
    \begin{minipage}[t]{0.49\textwidth}
    \captionsetup[subfloat]
    {}
    \centering
    \subfloat[Accuracy]{
    \label{figure:imdb_non_iid_label_flipping_a}
    \includegraphics[clip=true, trim=0 0 0 0, width=0.5\textwidth]{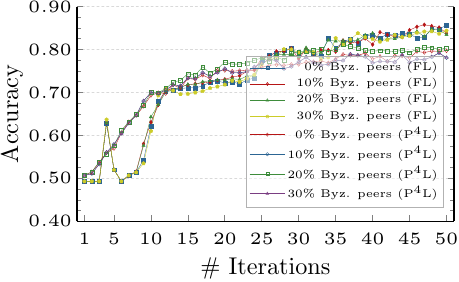}
    }
    \subfloat[AUC]{%
    \label{figure:imdb_non_iid_label_flipping_c}
    \includegraphics[clip=true, trim=0 0 0 0, width=0.5\textwidth]{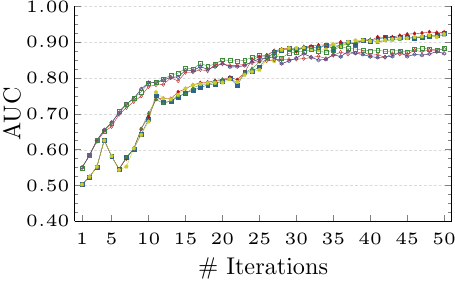}
    }
    \vspace{-0.75em}
    \caption{Effect of label flipping attack on Accuracy and AUC for \texttt{IMDB} (\texttt{sz}=3-10, \texttt{mrt}=50, \texttt{mrr}=50). 
    }\vspace{-0.4cm}
    \label{figure:imdb_non_iid_label_flipping}        
    \end{minipage}
\hfill
    \begin{minipage}[t]{0.49\textwidth}
    \captionsetup[subfloat]
    {}
    \centering
    \subfloat[Accuracy]{
    \label{figure:imdb_non_iid_byzantine_a}
    \includegraphics[clip=true, trim=0 0 0 0, width=0.5\textwidth]{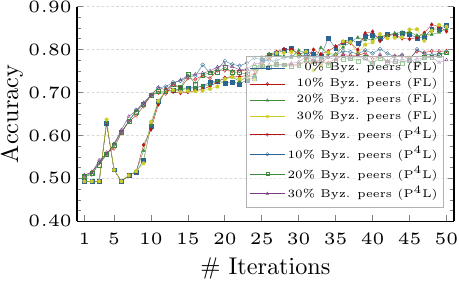}
    }
    \subfloat[AUC]{%
    \label{figure:imdb_non_iid_byzantine_c}
    \includegraphics[clip=true, trim=0 0 0 0, width=0.5\textwidth]{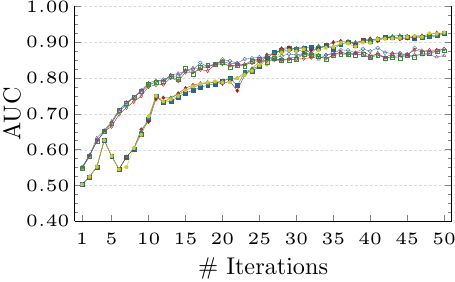}
    }
    \vspace{-0.75em}
    \caption{Effect of noisy peers on Accuracy and AUC for \texttt{IMDB} (\texttt{sz}=3-10, \texttt{mrt}=50, \texttt{mrr}=50). 
    }\vspace{-0.4cm}
    \label{figure:imdb_non_iid_byzantine}
    \end{minipage}
\end{figure*}

%% file: sections/6_implementation.tex
\section{Experimental Evaluation}
\label{sec:implementation}

\point{Prototype Implementation} To assess the effectiveness of our approach, we develop a prototype SDK of \toolname for the Android (v10 or greater) platform in Java. Specifically, we develop a simple Android app that demonstrates the basic operations of a peer when participating in a synergy: trains an ML model locally, scans the environment for available peers, connects to them to create a synergy, homomorphically adds and sends encrypted gradients, and finally decrypts the sum to perform the private aggregation. It is important to note that our approach does not impede the app's user experience, since it can work in the background and communicate with nearby peers without requiring any connection approval. Regarding Android permissions, our approach requires the user to allow access to their location, a permission that is necessary for searching and connecting to the available nearby peers.

Our implementation relies on the Nearby Connections API~\cite{nearby} and for the private aggregation part we implement a Homomorphic Encryption library that we developed based on the Paillier cryptosystem~\cite{paillierRust}. We chose the particular library because it allows homomorphic operations on packed ciphertexts (\ie ciphertexts that encrypt a vector of plaintext values instead of a single value). This way we improve the performance of our approach: large ciphertexts of traditional HE algorithms make homomorphic operations slow (as they have to manipulate these large ciphertexts) and application's throughput is suboptimal. Additionally, we use the gzip algorithm to compress our packed ciphertexts. We chose to implement our cryptographic library in Rust because (1) it provides cross-device capabilities (our core functionality is ready to be deployed in iOS), (2) its performance is optimal ~\cite{RustPerformance}. To handle big numbers, we use the high-performance Rust library of Arithmetic in Multiple Precision (RAMP)~\cite{ramp}. 

\point{Experimental Settings}
As an example ML use case, we consider an image classification problem that uses Transfer Learning i.e. a base model trained for a generic task is reused as the starting point for a model of another task. This type of learning is suitable for devices with limited h/w resources, e.g., smartphones. Here, as a base model we use MobileNetV2~\cite{sandler2018mobilenetv2}, pre-trained on ImageNet~\cite{5206848} with image size $224\times224$. As a head model, we use the same model discussed in Section~\ref{sssec:baselines_and_peer_models} ($\sim2,007,402$ trainable parameters). For model training and testing, we use \texttt{CIFAR-10}~\cite{krizhevsky2009learning}, equally split and distributed across all users. Similarly to other works~\cite{fls}, we use 20 samples per batch, 20 epochs and 200 samples per user.

\point{Experimental results}

We perform a series of measurements on the power discharge (in mAh) of the battery of three Google Pixel devices (Table~\ref{tab:specification-devices}), using the BatteryLab infrastructure~\cite{batterylab} that operationalises a \emph{Monsoon High Voltage Power Monitor}~\cite{monsoon}. Additionally, we measured the CPU utilization (in \%) and the end-to-end latency (in sec) of the smallest possible synergy. Figure~\ref{figure:performance} shows the average measurements across the different devices used, as a function of the various number of model parameters shared across the synergy. We see a linear increase of CPU utilization for the increasing number of exchanged model parameters, and a linear increase on the end-to-end latency (less than 4 sec for contemporary devices). The power discharge shows a similar trend, however the consumption is negligible (less than 3 mAh of discharge). 

\begin{table}[t]
\centering
\scriptsize
\begin{tabular}{ ll }
\toprule
 \bf Model & \bf Device Specification \\
  \midrule
 Pixel~3a & Octa-core (2x2.0 \& 6x1.7 GHz), 4GB RAM \\
 Pixel~4 & Octa-core (1x2.84, 3x2.42 \& 4x1.78 GHz), 4GB RAM \\
 Pixel~5 & Octa-core (1x2.4, 1x2.2 \& 6x1.8 GHz), 8GB RAM \\ \bottomrule
 \vspace{-0.3cm}
\end{tabular}
\caption{Test-bed Device Specification}
\label{tab:specification-devices}
\vspace{-0.9cm}
\end{table}

%% file: sections/7_conclusion.tex
\section{Conclusion}
\label{sec:conclusion}

Distributed ML approaches, and specifically FL, came to mitigate the profound issues of traditional ML systems on user privacy and bandwidth. Albeit of great importance, yet FL based approaches require centralized authority and maintenance that constitutes a single point of failure and bottleneck. In this paper, we put forth a new method of peer-to-peer learning (\toolname) that empowers users to engage in collaborative learning. Contrary to existing approaches that rely on external systems (\eg blockchain or IPFS), \toolname does not require any sort of infrastructure (no server or Internet connection). Instead, it leverages the proximity and cross-device communication capabilities of mobile devices, thus making it fault tolerant and highly scalable. \toolname uses strong cryptographic primitives to preserve confidentiality and utility of the shared gradients, while at the same time we eliminate information leakage attacks of malicious participants. We perform rigorous simulations to evaluate the effectiveness and resilience of our approach against several types of poisoning attacks, showing that it delivers global models that can exceed the performance of competitive baselines. Finally, we implement and evaluate the performance of \toolname, and we demonstrate its feasibility under ML models of various weights and devices, when it has negligible overheads on battery ($\ge$3mAh).